\documentclass[10pt,twocolumn,letterpaper]{article}
\usepackage{iccv}
\usepackage[accsupp]{axessibility}  % Improves PDF readability for those with disabilities.
\usepackage{authblk}
\newcommand\CoAuthorMark{\footnotemark[\arabic{footnote}]} % get the current value
\usepackage{times}
\usepackage{epsfig}
\usepackage{graphicx}
\usepackage{amsmath}
\usepackage{amssymb}
\usepackage{lipsum}
\usepackage{amsthm}
\usepackage{thmtools,thm-restate}
\usepackage{appendix}
\usepackage{pifont}
\usepackage{subcaption}
\usepackage{caption}
\newcommand{\cmark}{\ding{51}}%
\newcommand{\xmark}{\ding{55}}%
\usepackage{booktabs}
\usepackage{multirow}
\usepackage{multicol}
\usepackage[ruled,linesnumbered]{algorithm2e}
\SetKwInOut{Parameter}{Parameters}
\usepackage{etoolbox}
\AtBeginEnvironment{algorithm}{\SetArgSty{textrm}} 
\newcommand{\EXP}{\mathbb{E}}
\newcommand{\kpp}{$k$-means++}
\newcommand{\T}{\mathrm{T}}
\newcommand{\mS}{\mathcal{S}}

\DeclareMathOperator*{\argmax}{arg\,max}

\newtheorem{definition}{Definition}

\usepackage[breaklinks=true,bookmarks=false]{hyperref}

\iccvfinalcopy % *** Uncomment this line for the final submission

\def\iccvPaperID{****} % *** Enter the ICCV Paper ID here

\ificcvfinal\fi
\begin{document}

%%%%%%%%% TITLE
\title{DivAug: Plug-in Automated Data Augmentation with Explicit Diversity Maximization}

\author[1]{Zirui Liu\footnote{The first authors contribute equally to this paper.}}
\author[2]{Haifeng Jin\protect\CoAuthorMark}
\author[2]{Ting-Hsiang Wang}
\author[1]{Kaixiong Zhou}
\author[1]{Xia Hu}

\affil[1]{Rice Univeristy}
\affil[2]{Texas A\&M University}
\affil[ ]{\tt\small \{zl105, kz34, xia.hu\}@rice.edu}
\affil[ ]{\tt\small \{jin, thwang1231\}@tamu.edu}
% \author{Zirui Liu\footnote{The first authors contribute equally to this paper.} \qquad Haifeng Jin\footnotemark[\value{footnote}] \qquad  Ting-Hsiang Wang \qquad Kaixiong Zhou \qquad Xia Hu \\

% Department of Computer Science and Engineering, Texas A\&M University\\
% {\tt\small \{tradigrada, jin, thwang1231, zkxiong, xiahu\}@tamu.edu}
% % \and
% % Second Author\\
% % First line of institution2 address\\
% % {\tt\small secondauthor@i2.org}
% }

\maketitle

\ificcvfinal\thispagestyle{empty}\fi
\renewcommand{\thefootnote}{\fnsymbol{footnote}}
\footnotetext[1]{The first two authors contributed equally to this paper.}

%%%%%%%%% ABSTRACT

\begin{abstract}
Human-designed data augmentation strategies have been replaced by automatically learned augmentation policy in the past two years.
Specifically, recent work has empirically shown that the superior performance of the automated data augmentation methods stems from increasing the diversity of augmented data \cite{autoaug, randaug}. 
However, two factors regarding the diversity of augmented data are still missing: 1) the explicit definition (and thus measurement) of diversity and 2) the quantifiable relationship between diversity and its regularization effects.
To bridge this gap, we propose a diversity measure called Variance Diversity and theoretically show that the regularization effect of data augmentation is promised by Variance Diversity.
We validate in experiments that the relative gain from automated data augmentation in test accuracy is highly correlated to Variance Diversity.
An unsupervised sampling-based framework, \textbf{DivAug}, is designed to directly maximize Variance Diversity and hence strengthen the regularization effect.
Without requiring a separate search process,
the performance gain from DivAug is comparable with the state-of-the-art method with better efficiency.
Moreover, under the semi-supervised setting, our framework can further improve the performance of semi-supervised learning algorithms compared to RandAugment, making it highly applicable to real-world problems, where labeled data is scarce.
The code is available at \texttt{\url{https://github.com/warai-0toko/DivAug}}.
\end{abstract}
\vspace{-2mm}
\section{Introduction}
Data augmentation is a technique to create synthetic data from existing data with controlled perturbation.
For example, in the context of image recognition, data augmentation refers to applying image operations, \eg, cropping and flipping, to input images to generate augmented images, which have labels the same as their originals.
In practice, data augmentation has been widely used to improve the generalization in deep learning models and is thought to encourage model insensitivity towards data perturbation~\cite{krizhevsky2012imagenet, he2016deep, huang2017densely}.
Although data augmentation works well in practice, designing data augmentation strategies requires human expertise, and the strategy customized for one dataset often works poorly for another dataset.
Recent efforts have been dedicated to automating the design of augmentation strategies.
It has been shown that training models with a learned data augmentation policy may significantly improve test accuracy~\cite{fastautoaug, advaa, randaug, pba, fasterautoaug}.

However, we do not yet have a good theory to explain how data augmentation improves model generalization. Currently, the most well-known hypothesis is that data augmentation improves generalization by imposing a regularization effect: it regularizes models to give consistent outputs within the vicinity of the original data, where the vicinity of the original data is defined
as the space that contains all augmented data after applying operations that do not drastically alter image features~\cite{mixup, kernelda, wu2020generalization}.
Meanwhile, previous automated data augmentation works claim that the performance gain from applying learned augmentation policies arises from the  increase in diversity~\cite{autoaug, randaug, pba}.
However, the ``diversity'' in the claims remains a hand-waving concept: it is evaluated by the number of distinct sub-policies utilized during training or visually evaluated from a human perspective.
Without formally defining diversity and its relation to regularization, the augmentation strategies can only be evaluated indirectly by evaluating the models trained on the augmented data, which may cost thousands of GPU hours~\cite{autoaug}. 
It motivates us to explore the possibility of using an explicit diversity measure to quantify the regularization effect of the augmented data may have on the model. Thus, in this way we can directly maximize the diversity of the augmented data to strengthen the regularization effect to improve the generalization of the model.

\begin{figure*}
    \centering
    \includegraphics[scale=0.4]{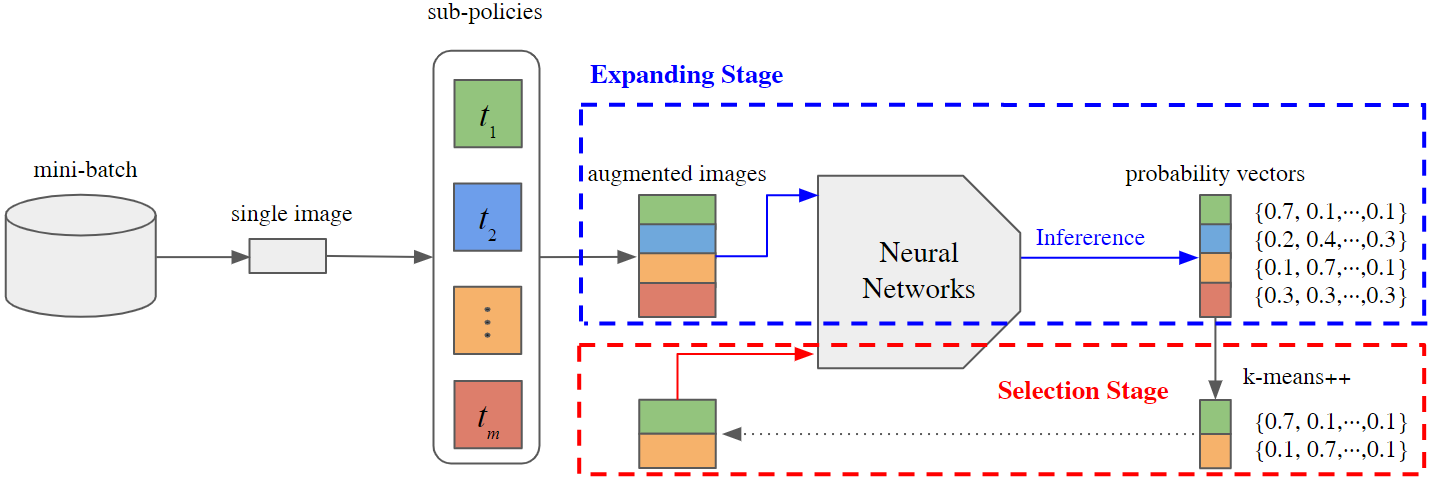}
    \caption{The DivAug framework overview.
    At the expanding stage, each data in the mini-batch is augmented by multiple randomly generated sub-polices.
    Notice the probability vectors of these augmented data are also obtained.
    At the selection stage, \kpp~seeding algorithm is used to sub-sample a subset of augmented data whose probability vectors are far apart from each other and thus diversifies the augmented data.
    Then the sampled data is used to train the model.}
    \vspace{-2mm}
    \label{fig:divaug_framwork}
\end{figure*}

To bridge the gap, in this paper we propose a new diversity measure, called Variance Diversity, to quantify the diversity of augmented data.
We show that the regularization effect of data augmentation is promised by Variance Diversity.
Our measure is motivated by the recent theoretical result that after applying augmented data to train the model, the loss implicitly contains a data-driven regularization term that is in proportion to the variance of probability vectors, where probability vectors are the outputs from models trained with the augmented data \cite{kernelda}.
Specifically,
we measure the diversity of a set of augmented data by the variance of their corresponding probability vectors. 
Based on the measure, we propose a plug-in automated data augmentation framework named \textbf{DivAug}, which can plug in the standard training process without requiring a separate search process. 
As illustrated in Figure \ref{fig:divaug_framwork}, the framework has two stages: the expanding stage, where we randomly generate several augmented data for each original input data,
and the selection stage, where we sub-sample a subset of augmented data and feed them to train the model.
Specifically, at the selection stage, for each image, we sub-sample a subset of augmented images with high diversity by applying the \kpp~seeding algorithm~\cite{kppseeding}, where the augmented data accompanied with probability vector which is far away from that of the original data is sampled with high probability.
Following the mathematical derivation, the regularization effect increases with the diversity of the augmented data. 
Consequently, the stronger regularization effect can lead to better model generalization, which is observed in terms of improved model performance.
Our main contributions can be summarized as follows:
\begin{itemize}
    \item We propose a new measure for quantifying the diversity of augmented data.
    We validate in our experiments that the relative gain in the accuracy of a model after applying data augmentation is highly correlated to our proposed measure.
    \item Based on the proposed measure, we design a sampling-based framework to explicitly maximize diversity.
    Without requiring a separate search process,
    the performance gain from DivAug is comparable to the state-of-the-art method with better efficiency.
    \item Our method is unsupervised and can plug in the standard training process. We show that our method  can further boost the performance of the semi-supervised learning algorithm, making it highly applicable to real-world problems, where labeled data is scarce.
\end{itemize}
\section{Related Work}
\noindent
Recently, AutoAugment (AA)~\cite{autoaug} has been proposed to automatically search for augmentation policies from a dataset.  
Specifically, AutoAugment utilizes a recurrent neural network (RNN) as the
controller to find the best policy in a separate search process on a small proxy task (smaller model size and dataset size).
Once the search process is over, the learned policies are transferred to the target task and fixed during the whole training process. 
These learned augmentation policies significantly improve the generalization of deep models~\cite{autoaug}. 
However, its search time is huge: 
it costs roughly 5,000 GPU hours to search for the best policies on a smaller dataset they call ``reduced CIFAR-10'', which consists of 4,000 randomly chosen images. 

\begin{table*}[hbt!]
\centering
\caption{Summary of automated data augmentation.}
\label{tab: summary of ada}
\begin{tabular}{|ccccc|} 
\hline
Method     & non-fixed & without the separate search process & unsupervised & without proxy tasks  \\ 
\hline
AA~\cite{autoaug}        & \xmark                 & \xmark            & \xmark                  & \xmark                 \\
Fast AA~\cite{fastautoaug}   & \xmark                 & \xmark            & \xmark                  & \cmark        \\
PBA~\cite{pba}       & \cmark                 & \xmark            & \xmark                  & \xmark                 \\
Adv. AA~\cite{advaa}   & \cmark        & \xmark            & \xmark                  & \cmark      \\
RA~\cite{randaug}        & \xmark        & \cmark   & \cmark         & \cmark        \\ 
\hline
\textbf{DivAug} (this paper)  & \cmark        & \cmark   & \cmark         & \cmark     \\
\hline
\end{tabular}
\vspace{-2mm}
\end{table*}

Most of the following works adopted the AutoAugment search space and formulation with improved optimization algorithms~\cite{advaa, fastautoaug, pba, fasterautoaug}.
Population-based augmentation (PBA)~\cite{pba} replaces the fixed policy with a dynamic schedule of policies evolving along with the training process. 
Fast AutoAugment (Fast AA)~\cite{fastautoaug} proposes a ``density match'' method to accelerate the search process and treats the augmented data as missing points in the training set. 
RandAugment (RA)~\cite{randaug} eliminates the separate search process by randomly applying augmentation sub-policies, which best resembles our work.
Adversarial AutoAugment (Adv. AA)~\cite{advaa} achieves state-of-the-art results by utilizing an RNN controller to learn policies that could generate augmented data with higher loss. 
As shown in Table \ref{tab: summary of ada}, we outline a general taxonomy of automated data augmentation methods, characterized by
four core properties. 
\textit{Non-fixed}: augmentation policies are dynamically changed along with the training process; 
\textit{without the separate search process}: methods do not require a separate search process; 
\textit{unsupervised}: methods do not require label information to find the  best policy;
and \textit{without proxy tasks}: methods perform the search directly on target tasks.

\section{Methodology}
In this section, we introduce the design and implementation of DivAug. 
First, we describe our search space in Section \ref{search space}. 
Then we mathematically show that after employing augmented data, the training loss implicitly contains a data-driven regularization term that is in proportion to the variance of probability vectors~(Section \ref{da analysis}). 
Subsequently, we propose to measure the diversity of a set of augmented data by the variance of their corresponding probability vectors. 
Based on the measure, we derive a sampling-based automated data augmentation method to explicitly maximize the diversity of augmented data~(Section \ref{divaug}).

% In this section, we present the proposed DivAug for generating augmented images with high diversity. We first analyze regularization effects of data augmentation. We found that regularization effects are related to variance of model's outputs to different operations~(section \ref{da analysis}). Subsequently, based on the theoretically insights, we derive a simple sampling based method to directly search sets of augmented images with high diversity on target datasets in the exponentially large search space~(section \ref{divaug}).

% describe the exponentially large search space of non-fixed augmentation policies.

\subsection{Search Space}
\label{search space}
\begin{figure}[hbt!]
    \centering
    \includegraphics[scale=0.35]{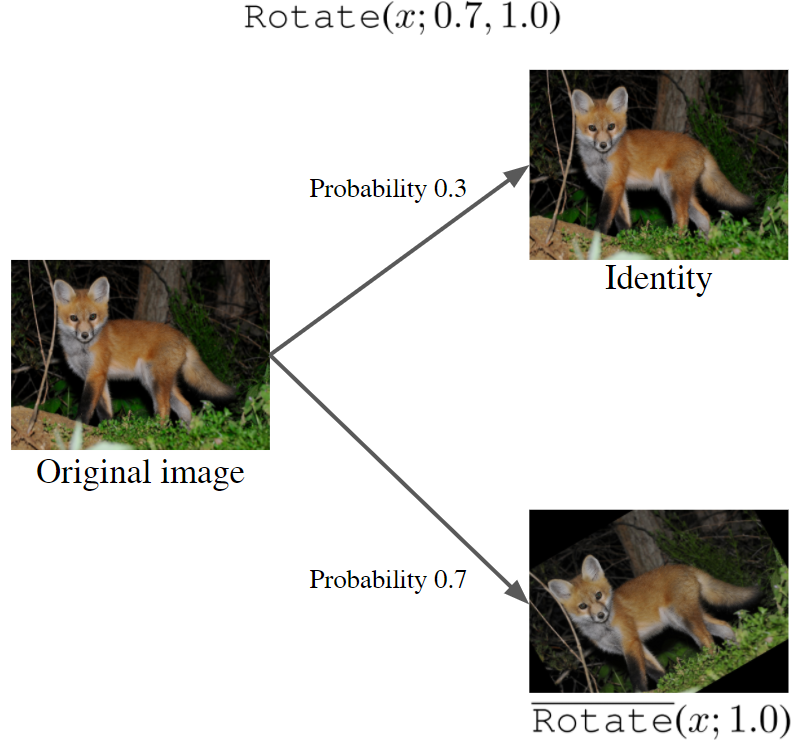}
    \caption{The schema of operation \texttt{Rotate}$(\cdot;0.7,1.0)$, where 1.0 is the normalized magnitude of the operation. Notice \texttt{Rotate}$(\cdot;0.7,1.0)$ denotes rotating the image by 30 degrees with the probability of 0.7.}
    \vspace{-2mm}
    \label{fig:illu of op}
\end{figure}

% \textcolor{red}{In this subsection, we introduce the design of our search space.} 
We adopt the basic structure of the well-designed search space introduced in AutoAugment~\cite{autoaug}. There are totally 16 image operations in our search space, including \texttt{Sharpness}, \texttt{ShearX/Y}, \texttt{TranslateX/Y}, \texttt{Rotate}, \texttt{AutoContrast}, \texttt{Invert}, \texttt{Equalize}, \texttt{Solarize}, \texttt{Posterize}, \texttt{Color}, \texttt{Brightness}, \texttt{Cutout}~\cite{cutout}, \texttt{Sample Pairing}~\cite{samplepairing}, and \texttt{Contrast}. Let $\mathcal{O}=\{\texttt{Sharpness},\cdots,\texttt{Contrast}\}$ be the set of all available operations. 
Each operation $\text{op}\in\mathcal{O}$ has two parameters: $p$, the probability of applying the operation; and $m$, the magnitude of the operation. 
To avoid creating confusion in notations, we use $\overline{\mathrm{op}}(\cdot;m)$ to represent image transformation specified by $\mathrm{op}$, with magnitude $m$. Given an image $x$, the operation $\mathrm{op}(x;p,m)$ is defined as:
\begin{equation}
    \mathrm{op}(x;p,m)=\left\{
    \begin{array}{ll}
         \overline{\mathrm{op}}(x;m), &\text{with probability}~p.\\
         x, &\text{with probability}~1-p.
    \end{array}
    \right.\nonumber
\end{equation}
% \textcolor{red}{Each operation comes with a default range of magnitudes to eliminate extreme image transformations.}
Each operation comes with a maximum range of magnitudes to avoid extreme image transformations.
For example, \texttt{Rotate} operation is only allowed to rotate images at most 30 degrees. 
The maximum range of magnitude for each operation is set to be the same as those reported in the AutoAugment. 
Meanwhile, we normalize the magnitude parameter $m$ to within $[0, 1]$, where $1$ stands for the maximum acceptable magnitude.
One example for illustrating the operation is shown in Figure \ref{fig:illu of op}.

In general, previous automated data augmentation methods search for the top augmentation policy, which is a set of five sub-policies, with each sub-policy consisting of two operations to be applied to the original images in sequence. 
Let $t$ be the sub-policy that consists of two consecutive operations, namely, $t(x)=\mathrm{op_2}(\mathrm{op_1}(x;p_1,m_1);p_2,m_2)$. 
For the sake of description convenience, we simplify the notation as $t:=\mathrm{op}_2\circ \mathrm{op}_1$.
Given the search space, previous automated data augmentation methods explore and rank the possible policy candidates in a separate search process.
Once the search process is over, the top five policies are collected to form a single final policy, which is a set containing 25 distinct sub-policies. 
The final policy is fixed throughout the training process. 
For each image in a mini-batch, only one sub-policy will be randomly selected to be applied~\cite{autoaug}.

% \textcolor{red}{Kx: too many first and second in this section. I totally re-organize this paragraph.}

However, the fixed policy may be sub-optimal due to the following two factors. First, there does not exist a sub-policy universally better than all other sub-policies throughout the training process~\cite{timematters, pba, advaa}.
For example, sub-policies that can reduce generalization error at the end of training is not necessarily a good sub-policy at the initial phase~\cite{affinity_diversity}. 
Second, the choices (hence diversity) of the augmented data is limited by the fixed set of unique sub-policies. From the above analysis, we design our search space similar to the AutoAugment's search space with two differences. 
First, inspired by Fast AutoAugment \cite{fastautoaug}, to introduce more stochasticity, we relax both the probability $p$ and magnitude $m$ as continuous parameters with value range $[0, 1]$. 
Second, the final policy in our search space is defined as the universal set that contains all the possible sub-policies. 
In contrast, the final policy in other work's search space is set to a fixed set of 25 unique sub-policies.
We note that RandAugment~\cite{randaug} samples the sub-policies uniformly over the search space similar to ours. The major distinctions in RandAugment are 1) the magnitude parameter $m$ is fixed discrete integer value, 2) the probability parameter $p$ is fixed to $\frac{1}{K}$. That means RandAugment \textit{always} applies operations on the original data\footnote{Although RA always applies operations, RA may keep the original image unchanged since its search space contains an \texttt{identity} operation.}.

\subsection{Regularization Effects of Data Augmentation}
\label{da analysis}
We derive the regularization effect of data augmentation following from the theoretical analysis in \cite{kernelda}.
We start by introducing the setting and notations of representation learning. 
Consider a neural network $f_\theta(x)$ parameterized by $\theta$. 
$f_\theta$ map the input $x$ into a vector representation $f_\theta(x)\in \mathbb{R}^D$ with $D$ output dimensions. 
We aim to minimize loss functions $l:\mathbb{R}^D\times\mathbb{R}\rightarrow \mathbb{R}$ over a dataset $\{(x_i, y_i)\}_{i=1}^{N}$, where $y_i \in \{1,\cdots D\}$. 
Let $\hat{p}(y|x)=$ \texttt{Softmax}($f_{\theta}(x)$) be the probability vector, where the \texttt{Softmax} function is used to normalize $f_\theta(x)$ into a probability distribution.
We denote the loss function to be minimized as $L=\sum_{i=1}^N L_i$, where $L_i=l(\hat{p}(y|x_i), y_i)$. 
We denote the gradient of $l$ with respect to the first argument as $l'\in\mathbb{R}^D$. 
Similarly, we use $l''\in\mathbb{R}^{D \times D}$ to represent the Hessian matrix of $l$ with respect to the first argument.
We use $t$ to represent the sub-policy, and $\T$ is the set of all available sub-policies. $x_i^t$ is the augmented data in the vicinity of $x_i$ obtained by applying $t$ to $x_i$. 
We use $\langle\cdot,\cdot\rangle$ to denote inner-product. 
For a set $\mathcal{S}$, we use $|\mathcal{S}|$ to represent its cardinality.
With these notations, after applying data augmentation, the new loss function becomes:

\begin{equation}
\label{eq:obj}
    \hat{L_i} = \EXP_{t \sim \T}[l(\hat{p}(y|x^t_i), y_i)].
\end{equation}

Suppose data augmentation does not significantly modify the feature map. Using the first order Taylor
approximation, we can expand Equation (\ref{eq:obj}) around point $\psi_i$:
\begin{align}
\label{eq:fo obj}
    \hat{L_i} \approx & l(\psi_i, y_i)+ \notag\\
&\EXP_{t \sim \T}[\langle \hat{p}(y|x^t_i)-\psi_i,l'(\psi_i,y_i)\rangle].
\end{align}
The second term in Equation (\ref{eq:fo obj}) can be cancelled by picking $\psi_i=\EXP_{t\sim \T}\hat{p}(y|x_i^t)$, \ie, $\psi_i$ is the averaged probability vector of all samples within the vicinity of $x_i$. 
If we further expand Equation (\ref{eq:obj}) around point $\psi_i=\EXP_{t\sim \T}\hat{p}(y|x_i^t)$ by considering the second order term, we have: 
\begin{align}
\label{eq:so obj}
\hat{L_i}\approx l(\psi_i, y_i)+\frac{1}{2}\EXP_{t\sim \T}[\Delta_i^\top l''(\psi_i,y_i)\Delta_i].
\end{align}
$\Delta_i :=\hat{p}(y|x^t_i)-\psi_i$ is the difference between the probability vector $\hat{p}(y|x_i^t)$ referring to the augmented data $x_i^t$, and the averaged probability vector $\psi_i$. 
The second term in Equation (\ref{eq:so obj}) is so called the ``data-driven regularization term'', which is exact the variance of the probability vector $\hat{p}(y|x_i^t)$, weighted by $l''(\psi_i,y_i)$. 
That means employing augmented data imposes a regularization effect by implicitly controlling the variance of model's outputs.

% Further, in the context of image recognition, Proposition 1 below shows that this regularization effect is label-independent, \ie, the second term of (\ref{eq:so obj}) does not involve any label information.

% \begin{proposition}
% \label{prop indenpend}
% If the loss function $l$ is cross-entropy loss, the second term in (\ref{eq:so obj}) is independent of label $y$.
% \end{proposition}

% \begin{proof}
% We prove Proposition \ref{prop indenpend}
%  by showing $l''(\psi,y)$ is independent of $y$ if $l$ is cross-entropy loss. 

% We denote $\mathbf{W}^\top\psi$ as $\tilde{p} \in \mathbb{R}^D$ and $\tilde{p}_i$ to represent the $i$-th entry of vector $\tilde{p}_i$, respectively. 
% We denote \texttt{Softmax}($\tilde{p}$) as $\hat{p}$, with $\hat{p}_i=\log(\frac{e^{\tilde{p}_i}}{\sum_{j=1}^D e^{\tilde{p}_j}})$.
% We use $l_{\CE}$ to represent cross-entropy loss.  
% Let $\hat{y}$ be the one-hot vector corresponding to the label $y$.
% With these notations, we have:
% \begin{equation}
%     l_{\CE}(\tilde{p},y)=\log(\sum_{j=1}^D
%     e^{\tilde{p}_j})- \tilde{p}_y
% \end{equation}
% It is easy to check that $l'=\frac{\partial l_{\CE}}{\partial \tilde{p}}=\hat{p}-\hat{y}$, and $l''=\frac{\partial l'}{\partial \tilde{p}}$ is independent of $y$.
% \end{proof}

\subsection{The DivAug Framework}
\label{divaug}
To establish the relationship between the diversity of augmented data and their regularization effect, we propose a new diversity measure, called Variance Diversity, for the augmented data whose regularization effect can be quantified.
Based on this, we derive a sampling-based framework that explicitly maximizes the Variance Diversity of the augmented data.
\vspace{-2mm}
% We introduce our proposed framework to improve generalization of models in this subsection. 
% The key idea of our proposed framework is to increase the regularization effect by training models with multiple diversely augmented data.

\begin{algorithm*}[hbt!]
\caption{DivAug}
\label{algo:divaug}
\KwIn{input image $x$; model $f_{\theta}$; all possible operations ~$\mathcal{O}$=$\{$\texttt{Sharpness},$\cdots$, \texttt{Contrast}$\}$}
\Parameter{the number of augmented images per input image $E$; the number of selected augmented images per input image used for training $S$
}
\KwOut{$\mS:=$ a set of $S$ augmented images of input image $x$}
\For{$j=1,\cdots, E$}{
    Sample operations $\mathrm{op_1, op_2}\sim\mathcal{O}$ uniformly at random \\
    $p_1\sim \mathrm{Uniform}(0, 1); p_2\sim \mathrm{Uniform}(0, 1)$ \\
    $m_1\sim \mathrm{Uniform}(0, 1); m_2\sim \mathrm{Uniform}(0, 1)$ \\
    Get sub-policy $t_j:=\mathrm{op_1}(\cdot~; p_1, m_1)\circ \mathrm{op_2}(\cdot~; p_2, m_2)$\\
    Generate $x^{t_j}=t(x)$\\
    Compute $\hat{p}(y|x^{t_j})$= \texttt{Softmax}$(f_{\theta}(x^{t_j}))$}
Generate a set of augmented images $\mS$ of size $S$, which is a random subset of $\{x^{t_j},j=1,\cdots,E\}$, using \kpp~ seeding algorithm on $\{\hat{p}(y|x^{t_j}):j=1,\cdots,E\}$\\
\Return{$\mS$}
\end{algorithm*}

\subsubsection{Diversity Measure of Augmented Data}
\label{Diversity of Augmented Data}
We start by proposing a new diversity measure for augmented data, whose regularization effect can be quantified. 

% Then we discuss how our proposed diversity measure is related to other existing measures.

\begin{figure}[hbt!]
    \centering
    \includegraphics[scale=0.21]{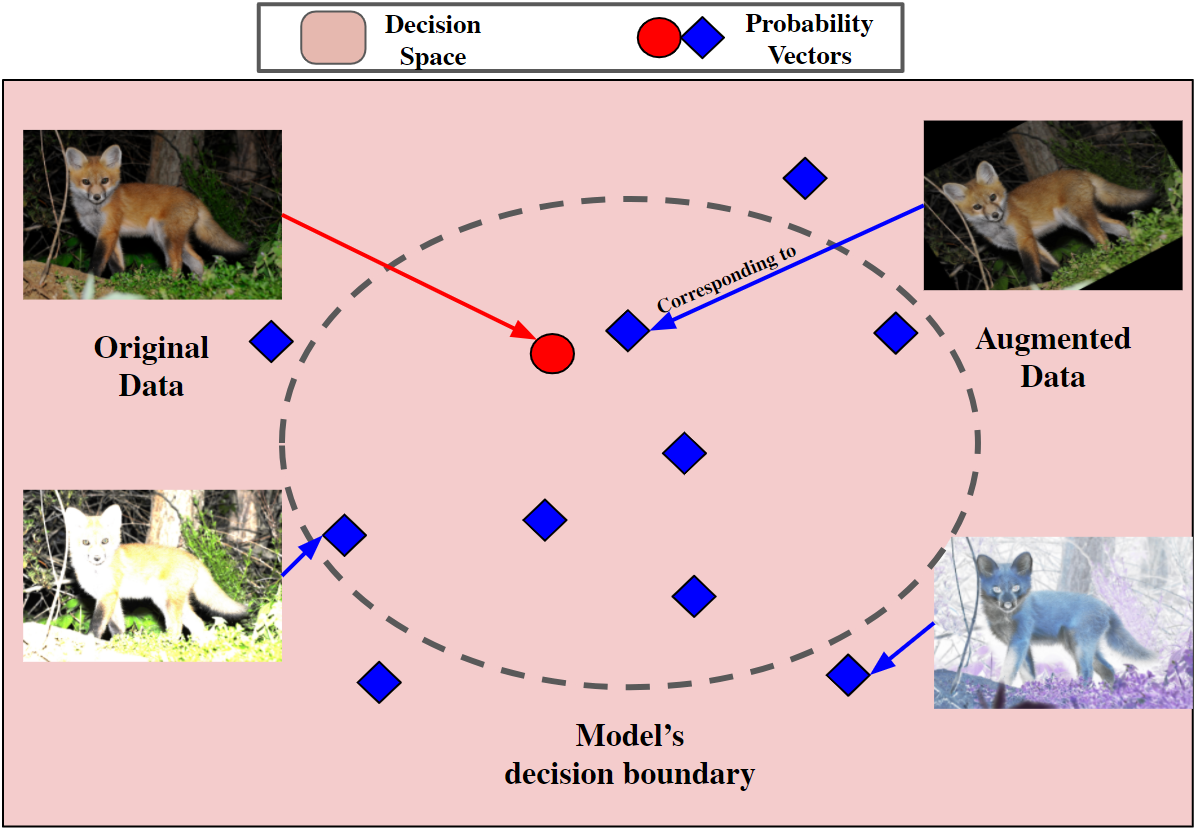}
    \caption{An example to illustrate the diversity between augmented data. 
    DivAug explicitly looks for augmented data whose corresponding probability vectors are far away from each other in the decision space.}
    \label{fig:illu of diversity}
    \vspace{-1em}
\end{figure}

From Equation (\ref{eq:so obj}), after training models on augmented data, a data-driven regularization term can be decomposed from the loss function. 
From above, 
we quantify the diversity of a set of augmented data by the variance of their corresponding probability vectors. 
Formally, given a model $f_{\theta}$, for a set of augmented data $\mS=\{x^{t_j}\}_{j=1}^{S}$, where $x^{t_j}$ is generated from the same original data $x$ by applying different sub-policy $t_j$, we define the diversity of $\mS$ as:
\begin{equation}
\label{eq: def diversity}
    % \mathcal{D}(\mathcal{S})=\EXP_{x^t\in\mathcal{S}}\Delta^\top \Delta.
    \mathcal{D}(\mathcal{S})=\EXP_{x^{t_j}\in\mathcal{S}}\Delta^\top \Delta.
\end{equation}
$\hat{p}(y|x^{t_j}):=$ \texttt{Softmax}($f_{\theta}(x^{t_j})$) is the probability vector corresponding to $x^{t_j}$, and $\Delta = \hat{p}(y|x^{t_j}) - \EXP_{x^{t_j}\in\mathcal{S}}\hat{p}(y|x^{t_j})$.
If CrossEntropy is used as the loss function, then the Hessian matrix $l''(\psi_i,y_i)$  is a diagonal matrix, where the elements on the diagonal are all zero, except for the one corresponding to the true label. 
This implies that under the supervised setting, only the variance of the probability associated with the true label will be penalized.
We can extend this penalty effect to unsupervised domain by setting $l''(\psi_i,y_i)$ in Equation (\ref{eq:so obj}) as the identity matrix. 
In this way, Equation (\ref{eq:so obj}) penalizes the variance of the probability associated with any class. We note that this is essentially the consistency regularization, which is one of the key techniques in semi-supervised learning and self-supervised learning, which encourages the model to produce similar probability vectors when the input data is perturbed  by noise \cite{uda, mixmatch}.
Moreover, if $l''(\psi_i,y_i)$ in Equation (\ref{eq:so obj}) is set as the identity matrix,
the diversity of augmented data is exact the data-driven regularization term in Equation (\ref{eq:so obj}).

According to Equation (\ref{eq: def diversity}), we name our diversity measure ``Variance Diversity''. 
We note that this is a unsupervised model-specific measure, which depends only on the model prediction without involving any label information.
Intuitively, as illustrated in Figure \ref{fig:illu of diversity}, 
if a set of augmented data has large Variance Diversity, that means their corresponding probability vectors are far away from each other.
Therefore, it is harder for models to give consistent predictions for diversely augmented data.
This forces the models to generalize over the vicinity of original data.

\subsubsection{Design of DivAug}
According to the definition of Variance Diversity and Equation (\ref{eq:so obj}), the increase of Variance Diversity directly strengthens the regularization effect of augmented data.
Based on this insight, our DivAug framework generates a set of diversely augmented data and minimizes the loss over them. 
Specifically, DivAug consists of two stages: the expanding stage and the selection stage. 
At the expanding stage, for each original data $x_i$, we first randomly generate a set of sub-policies $\{t_j\}_{j=1}^{E}$, where $\{x_i^{t_j}\}_{j=1}^{E}$ are the set of augmented data $x_i^{t_j}$ corresponding to $t_j$. 
The second stage is the selection stage, 
where we sub-sample a subset of augmented data $\mathcal{S}_i\subset \{x_i^{t_j}\}_{j=1}^{E}$, where $|\mathcal{S}_i|=S<E$. 
Then we feed the selected augmented data to the model.
Our DivAug framework is illustrated in Figure \ref{fig:divaug_framwork}.
Formally, with the notations introduced in Section \ref{da analysis} and Section \ref{Diversity of Augmented Data}, given $f_{\theta}$, we minimize the following objective:
\begin{align}
    \min_{\theta}~~~\frac{1}{N}&\sum_{i=1}^N[\frac{1}{S}\sum_{x_i^{t}\in \mS_i} l(\hat{p}(y|x_i^{t}), y)], \\
    \label{eq: var of augmented data}
    \mathrm{s.t.}~~~&\mS_i = \argmax_{\substack{\mS_i\subset{\{{x_i^{t_j}\}}^E_{j=1},}\\|\mS_i|=S}}
    \EXP_{x_i^{t_j}\in \mS_i} \Delta_i^\top\Delta_i.
\end{align}

where $\Delta_i=\hat{p}(y|x_i^{t_j})-\EXP_{x_i^{t_j}\in\mS_i}\hat{p}(y|x_i^{t_j})$.
From Equation (\ref{eq: var of augmented data}), we target at selecting a subset of augmented data $\mathcal{S}_i$, 
whose corresponding probability vectors have maximum variance. 
Unfortunately, getting the solution of Equation (\ref{eq: var of augmented data}) poses a significant computational hurdle.
Instead of computing the optimal solution, we efficiently sample $\mathcal{S}_i$ with the \kpp~seeding algorithm \cite{kppseeding}, which is originally made to generate a good initialization for $k$-means clustering. 
\kpp~seeding selects centroids by iteratively sampling points in proportion to
their squared distances from the closest centroid that has been chosen.
Here, we define the distance between a pair of probability vector as their Euclidean distance. 
Therefore, \kpp~ samples a subset of augmented data where their probability vectors are far apart from each other, which practically leads to a large Variance Diversity.
For more details, the \kpp~seeding algorithm is shown in Algorithm \ref{algo: kpp seeding} in the Appendix \ref{app: kpp seeding}. 
We show the algorithm of DivAug in Algorithm \ref{algo:divaug} and remark that the operation is randomly generated.
There are two hyperparameters in Algorithm \ref{algo:divaug}. Namely, the number of augmented images per input image $E$, and the number of selected augmented images per input image used for training $S$.
Moreover, the two hyperparameters $S$ and $E$ do not need to be tuned on proxy tasks and can be chosen according to available computation resources.
Similar to RandAugment, DivAug is a sampling-based method that does not require a separate search process.
Note that there is no label information involved in Algorithm \ref{algo:divaug}, which means DivAug is suitable for both semi-supervised learning and supervised learning.

\section{Experiments}
Our experiments aim to answer the following research questions:
\begin{itemize}
    \item \textbf{RQ1}.~What is the effect of Variance Diversity on model generalization?
    \item \textbf{RQ2}.~How effective is the proposed DivAug compared with other automated data augmentation methods under the supervised settings?
    \item \textbf{RQ3}.~How well does DivAug improve the performance of semi-supervised learning algorithms?
\end{itemize}

\subsection{Experimental Settings}
\label{exp: exp setting}
\noindent
Below, we first introduce the datasets and the default augmentation method for them. Then, we will introduce the hyperparameter setting of Divaug ($S$ and $E$ in Algorithm \ref{algo:divaug}) , and the baseline methods for comparison.

We adopt four benchmark datasets for evaluating our proposed method: CIFAR-10, CIFAR-100, SVHN and ImageNet.
These four datasets
are processed based on the way and codes provided in \cite{autoaug}.
The basic statistics of these four datasets and the default data augmentation for them are summarized in Appendix.
For DivAug, we set $E=8$ and $S=4$ for the experiments in Section \ref{exp: rq1} and \ref{exp: supversied}, excluding the ImageNet experiment.
For ImageNet, we set $E=4$ and $S=2$ due to limited resources.
For the semi-supervised learning experiment, we set $E=4$ and $S=2$.
We did not tune these two hyperparameters, and we choose them mainly according to the available GPU memory.

The methods for comparison are as below:
We compare Algorithm \ref{algo:divaug} with AutoAugment (AA) \cite{autoaug}, Fast AutoAugment (Fast AA) \cite{fastautoaug}, Population Based Augmentation (PBA) \cite{pba}, RandAugment (RA) \cite{randaug}, and Adversarial AutoAugment (Adv. AA) \cite{advaa}. 
For each image, the augmentation policy proposed by different methods and the default augmentation are applied in sequence.w

\subsection{Correlation Between Variance Diversity and Generalization}
\label{exp: rq1}

To answer \textbf{RQ1}, we calculate the Variance Diversity of augmented data generated by AA, Fast AA, RA, the default augmentation introduced in Section \ref{exp: exp setting}, and DivAug\footnote{We do not include Adv. AA because the official code is not released. For PBA, the official code is based on Ray and hard to migrate our codebase for a fair comparison.}. Then, we report the test accuracy of models trained on augmented data generated by different methods.

\begin{figure}[hbt!]
\centering
    \includegraphics[scale=0.4]{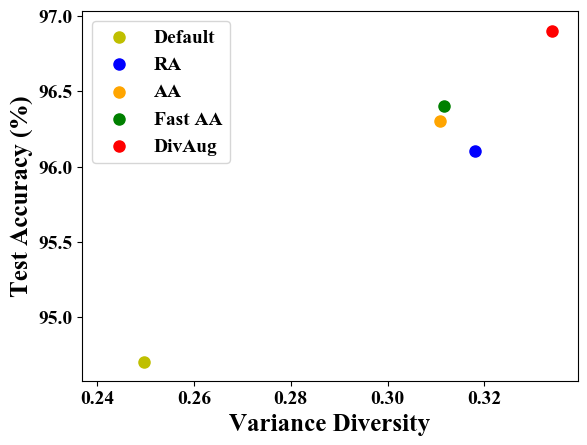}
    \caption{
    \textbf{
    The performance gain is positively correlated to Variance Diversity.}
    In general, almost all points lies near the diagonal, and the relative gain in test accuracy increases with larger Variance Diversity. 
    }
    \vspace{-1em}
    \label{fig: aff var corr}
\end{figure}

Because Variance Diversity is an unsupervised, model-specific measure, for a fair comparison, 
we first train a Wide-ResNet-40-2 model on CIFAR-10 without applying any data augmentation methods. Then we use it as the $f_{\theta}$ in Equation (\ref{eq: def diversity}) to evaluate all different automated data augmentation methods.
To verify the correlation between generalization and Variance Diversity,
we calculate the Variance Diversity of augmented data as follows:
for each image in the training set, an automated augmentation method is used to randomly generate four augmented images. Then we calculate the Variance Diversity of these four images according to Equation (\ref{eq: def diversity}). 
We report the averaged Variance Diversity over the entire training set in Figure \ref{fig: aff var corr}. 

Figure \ref{fig: aff var corr} demonstrates the performance gain and Variance Diversity are positively correlated (the detailed test accuracy is shown in the first row of Table \ref{tab: supervised res}).
As shown in the figure, all automated data augmentation methods could improve the Variance Diversity of augmented data over the default augmentation.
Specifically, AA and Fast AA has small Variance Diversity. 
It makes sense because both of them try to minimize the distribution shift of the augmented data from the original distribution. 
For example, Fast AA treats the augmented data as the missing point in the training set.
As a result, for CIFAR-10, all of the reported sub-policy proposed by AA and Fast AA do not contain the counter-intuitive operation \texttt{SamplePair} \cite{autoaug, fastautoaug}, which limits the Variance Diversity of the augmented data generated by them.
In contrast, DivAug has the largest Variance Diversity because it tries to explicitly maximize the Variance Diversity of the augmented data. Notice
RA has larger Variance Diversity compared to AA and Fast AA. This might be a result of RA randomly sample operations. As a result, RA samples more distinct sub-policies than AA and Fast AA do and leads to larger diversity. 
Here we remark that although RA has larger Variance Diversity compared to AA and Fast AA, the model's relative gain in accuracy is smaller compared to those of AA and Fast AA.
We provided a detailed analysis in the Appendix \ref{app: detailed analysis of the correlation}.

We also present a simple case study in Appendix \ref{app: ablation study} Figure \ref{fig: rotate_case_study}, where DivAug's candidate images are obtained by only applying the single transform \texttt{Rotate} with fixed probability parameter $p$ (the magnitude parameter remains random). As shown in Figure \ref{fig: rotate_case_study}, Variance Diversity and generalization are generally correlated.

\subsection{The Effectiveness of DivAug Under the Supervised Settings}
\label{exp: supversied}
\begin{table*}
\centering
\caption{Test accuracy (\%) on CIFAR-10 and CIFAR-100. For ImageNet, we report the validation accuracy (\%). We compare our method with the default data augmentation (Baseline), AA, Fast AA, PBA, RA, and Adv. AA. Our results are averaged over four trials except ImageNet.}
\label{tab: supervised res}
\begin{tabular}{|cc|cccccc|c|} 
\hline
Dataset                                        & Model                                                                     & Baseline                 & AA                       & Fast AA                  & PBA                   & RA                       & Adv. AA                   & DivAug                     \\ 
\hline
\multicolumn{1}{|l}{\multirow{4}{*}{CIFAR-10}} & \textcolor[rgb]{0.2,0.2,0.2}{Wide-ResNet-40-2}                            & 94.7                     & 96.3                     & 96.4                     & -                     & 96.1                     & -                         & \textbf{96.9}$\pm$.1       \\
\multicolumn{1}{|l}{}                          & \textcolor[rgb]{0.2,0.2,0.2}{Wide-ResNet-28-10}                           & 96.1                     & 97.4                     & 97.3                     & 97.4                  & 97.3                     & \textbf{98.1}             & \textbf{98.1}$\pm$.1       \\
\multicolumn{1}{|l}{}                          & \textcolor[rgb]{0.2,0.2,0.2}{Shake-Shake (26 2x96d)}                      & 97.1                     & 98.0                     & 98.0                     & 98.0                  & 98.0                     & \textbf{98.1}             & \textbf{98.1}$\pm$.1       \\
\multicolumn{1}{|l}{}                          & \multicolumn{1}{l|}{\textcolor[rgb]{0.2,0.2,0.2}{PyramidNet+ShakeDrop } } & 97.3                     & 98.5                     & 98.3                     & 98.5                  & 98.5                     & \textbf{98.6}             & 98.5$\pm$.1                \\ 
\hline
\multirow{3}{*}{CIFAR-100}                     & \textcolor[rgb]{0.2,0.2,0.2}{Wide-ResNet-40-2}                            & 74.0                     & 79.3                     & 79.4                     & -                     & -                        & -                         & \textbf{81.3}$\pm$.3       \\
                                               & \textcolor[rgb]{0.2,0.2,0.2}{Wide-ResNet-28-10}                           & 81.2                     & 82.9                     & 82.7                     & 83.3                  & 83.3                     & \textbf{84.5}             & 84.2$\pm$.2                \\
                                               & \textcolor[rgb]{0.2,0.2,0.2}{Shake-Shake (26 2x96d)}                      & 82.9                     & 85.7                     & 85.1                     & 84.7                  & -                        & \textbf{85.9}             & 85.3$\pm$.2                \\ 
\hline
SVHN                                       & Wide-ResNet-28-10                                                                 & 96.9                    & 98.1                    & -                     & -                     & \textbf{98.3}                     & -            & \textbf{98.3}$\pm$.3                       \\ 
\hline
ImageNet                                       & ResNet-50                                                                 & 76.3                     & 77.6                     & 77.6                     & -                     & 77.6                     & \textbf{79.4}             & 78.0     \\
\hline
\end{tabular}
\end{table*}

The main propose of automated data augmentation is to further improve the generalization of models over traditional data augmentation techniques.
To answer \textbf{RQ2}, we compare our proposed method with several baselines under the supervised learning settings. 

\subsubsection{Experiment on CIFAR-10 and CIFAR-100}
Following \cite{autoaug, fastautoaug, randaug},
we evaluate our proposed method with the following
models: Wide-ResNet-28-10, Wide-ResNet-40-2 \cite{wideresnet}, Shake-Shake (26 2x96d) \cite{shakeshake}, and PyramidNet+ShakeDrop \cite{shakedrop, pyramidnet}. 
The details of hyperparameters are shown in Appendix Table \ref{tab: hp supervised}.

\textbf{CIFAR-10 Results:}
In Table \ref{tab: supervised res}, 
we report the test accuracy of these models. 
For all of these models, our proposed method can achieve better performance compared to previous methods. We achieve $0.7\%$, $0.8\%$, $0.7\%$, $0.8\%$ improvement on Wide-ResNet-28-10 compared to AA, Fast AA, PBA and RA, respectively.
Overall, DivAug significantly improves the performances over baselines
while achieves comparable performances to those of Adv. AA.

\begin{figure}[hbt!]
\centering
    \begin{subfigure}[b]{.4345\linewidth}
      \includegraphics[width=1\linewidth]{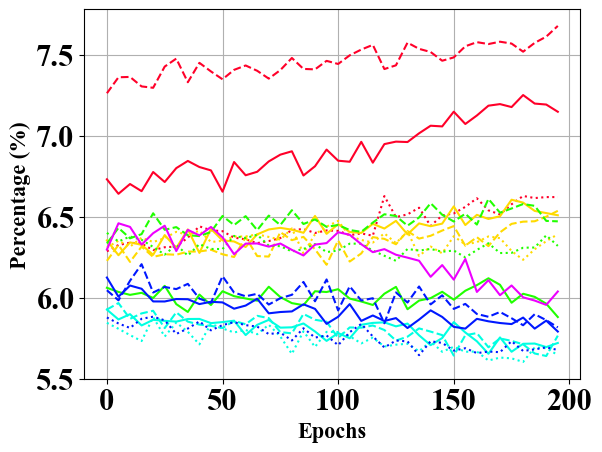}
      \label{fig:sub-first}
    \end{subfigure}
    \begin{subfigure}[b]{.55\linewidth}
    \includegraphics[width=1\linewidth]{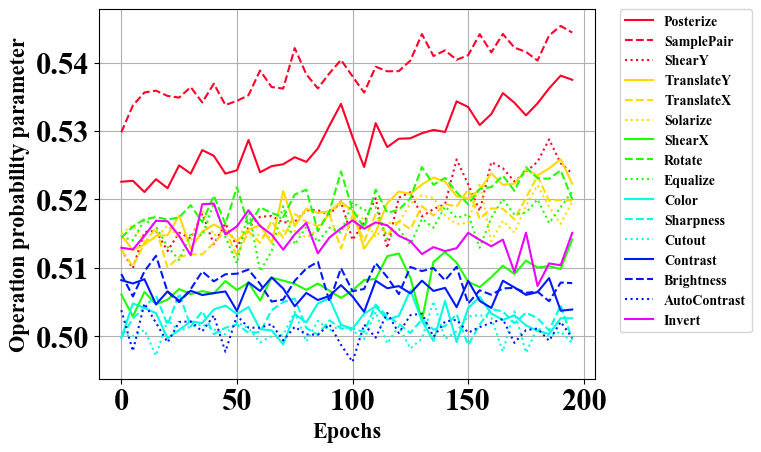}
      \label{fig:sub-first}
    \end{subfigure}
    \vspace{-2.5em}
    \caption{
    \textbf{The distribution of selected sub-policies evolves along with the training process.}
    (a) The statistics of sub-policies selected by DivAug.
    (b) The averaged probability of applying operations in sub-policies selected by DivAug.}
    \vspace{-2mm}
    \label{fig: stat over epoch}
\end{figure}

\textbf{The effect of \kpp~:}
To check the effect of \kpp~in DivAug, we compare the performance of Wide-ResNet-28-10 with DivAug and that with the random baseline in Appendix \ref{app: ablation study} Table \ref{tab: random_baseline}.
The random baseline here randomly picks $S$ augmented images from $E$ candidates for training.
Also, the magnitude $m$ and probability $p$ are also randomly picked.
As shown in Table \ref{tab: random_baseline}, DivAug is significantly better than the random baseline.
Moreover, to understand the effect of \kpp~and how DivAug improves the test accuracy over RA,
we further visualize the distribution of sub-policies selected by DivAug with Wide-ResNet-40-2 on CIFAR-10 over the training process.
As shown in Figure \ref{fig: stat over epoch}, we observe that the percentages of some operations picked from the sampled sub-policies, such as \texttt{TranslateY}, \texttt{ShearY}, \texttt{Posterize}, and \texttt{SampleParing}, gradually increase along with the training process. 
In contrast, some color-based operation, such as \texttt{Invert}, \texttt{Brightness}, \texttt{AutoContrast}, and \texttt{Color}, gradually decrease along with the training process.
This behavior is consistent with the discovery that there does not exist an operation beating all other operations throughout the training process \cite{pba, advaa}.
Also, the average probability of applying operations in the selected sub-policies slowly increases with the training process. That means DivAug tends to mildly shift the distribution of augmented images away from the original one over the training process.
From above, it suggests that the sub-policies selected by DivAug evolve throughout the training process.
\vspace{-1em}

\textbf{Training Efficiency Analysis:}
DivAug is estimated to be significantly faster than Adv. AA for the following reasons.
Following the time cost metric in \cite{wu2020generalization}, we estimate the inference cost (see Algorithm \ref{algo:divaug} line 7) equals half of the training cost.
Under the setting of $E=8$ and $S=4$, DivAug additionally generates four times more augmented data for training. 
In contrast,
Adv. AA needs to generate eight times more augmented data to achieve the results reported in Table \ref{tab: supervised res}.
Moreover, it also needs a separate phase to search for the best policy.
Although the search time for Adv. AA is not reported in \cite{advaa}.
The estimated costs are summarized in Table \ref{tab: cost of divaugment}.

\begin{table}[hbt!]
\centering
\caption{Comparison of the total cost of DivAug and Adv. AA on CIFAR-10 relative to RA. The training cost of Adv. AA is cited from \cite{advaa}.}
\label{tab: cost of divaugment}
\begin{tabular}{c|ccc}
\hline
         & RA & ~Adv. AA & DivAug  \\ 
\hline
Training($\times$) & 1.0   &    8.0 + Search Cost     &      4.5   \\
\hline
\end{tabular}
\end{table}

\textbf{CIFAR-100 Results:}
As shown in Table \ref{tab: supervised res}, DivAug generally achieves non-trivial performance gain over all other methods excluding Adv. AA.
However, we note that DivAug does not require label information or a separate search process. 
Also, DivAug is significantly faster than Adv. AA.
% \vspace{-2em}

\subsubsection{Experiment on ImageNet}
Following \cite{autoaug, fastautoaug, randaug},
we select ResNet-50 \cite{he2016deep} to evaluate our proposed method.
The details of the hyperparameters are shown in Appendix Table \ref{tab: hp supervised}.
As shown in Table \ref{tab: supervised res}, DivAug outperforms other baselines except Adv. AA.
We remark that due to the limited resources, the two hyperparameters in Algorithm \ref{algo:divaug} are set to $E=4$ and $S=2$, respectively.
The performance gain from DivAug is expected to be further improved with larger $E$ and $S$.

\subsection{The Effectiveness of DivAug Under the Semi-Supervised Setting}
\label{exp: ssl}

% Semi-supervised learning \cite{chapelle2009semi} (SSL) has been shown to be an effective framework to alleviate the reliance on labeled datasets by leveraging unlabeled data.
One of the key techniques in semi-supervised learning \cite{chapelle2009semi} (SSL) is consistency regularization, which encourages the model to produce similar probability vectors when the input data is perturbed by noise.
It has been proven that the augmented data produced by state-of-the-art automated methods can serve as a superior source of noise under the consistency regularization framework \cite{uda, fixmatch}.
Specifically, UDA \cite{uda} utilizes RA as the source of perturbation and achieves non-trivial performance gain. Also, it has been theoretically shown that the success of UDA stems from the diversity of augmented data generated by RA \cite{uda}.

However, most automated data augmentation methods require label information to search for the best policy. 
Thus, this prerequisite limits their application in SSL. 
In contrast, our proposed method is suitable for SSL because it is unsupervised and tries to explicitly maximize diversity.
This leads to the following question: can SSL benefit from our proposed DivAug (\textbf{RQ3})?
To answer this question, following UDA, we change the source of perturbation from RA to DivAug (detailed hyperparameters are shown in the Appendix).
Here, we report the averaged results over four random trials. 
As shown in Table \ref{tab: ssl results}, DivAug can further boost the performance of UDA under different settings.
Moreover, the performance gap grows larger when there is less labeled data available.
This might be because, when there is limited labeled data, the regularization effect brought by diversity plays a much bigger role in model performance.

\vspace{-2mm}
\begin{table}[hbt!]
\centering
\caption{Error rate (\%) comparison on CIFAR-10 with 1000, 2000, and 4000 labeled data.
The architecture is Wide-ResNet-28-2.
For fair comparison, we reproduced the UDA(RA)$^*$ result by ourselves using the same codebase.}
\label{tab: ssl results}
\begin{tabular}{cccc} 
\hline
\multirow{2}{*}{Methods} & \multicolumn{3}{c}{CIFAR-10}                                            \\ 
\cline{2-4}
                         & 1000  & 2000        & 4000\\ 
\hline
UDA(RA)$^{*}$            & 7.37$\pm$0.15         & 6.50~$\pm$0.14  & 5.44$\pm$0.15             \\
UDA(DivAug)              & \textbf{6.94} $\pm$0.12  & \textbf{6.26}$\pm$0.15   & \textbf{5.40}$\pm$0.12\\
\hline
\end{tabular}
\end{table}

\vspace{-1em}
\section{Conclusion}
In this work, we propose a new diversity measure called Variance Diversity by investigating the regularization effect of data augmentation. 
We validate in experiments that the performance gain from automated data augmentation is highly correlated to Variance Diversity. 
Based on this measure, we derive the DivAug framework to explicitly maximize Variance Diversity during training. 
We demonstrate our proposed method has the practical utility of achieving better performance without the need to search for top policies in a separate phase.
Therefore, DivAug can benefit both the supervised tasks and the semi-supervised tasks.

{\small
\bibliographystyle{ieee_fullname}
\bibliography{reference}

\begin{thebibliography}{10}\itemsep=-1pt

\bibitem{kppseeding}
David Arthur and Sergei Vassilvitskii.
\newblock k-means++ the advantages of careful seeding.
\newblock In {\em Proceedings of the eighteenth annual ACM-SIAM symposium on
  Discrete algorithms}, pages 1027--1035, 2007.

\bibitem{mixmatch}
David Berthelot, Nicholas Carlini, Ian Goodfellow, Nicolas Papernot, Avital
  Oliver, and Colin~A Raffel.
\newblock Mixmatch: A holistic approach to semi-supervised learning.
\newblock In {\em Advances in Neural Information Processing Systems}, pages
  5049--5059, 2019.

\bibitem{chapelle2009semi}
Olivier Chapelle, Bernhard Scholkopf, and Alexander Zien.
\newblock Semi-supervised learning (chapelle, o. et al., eds.; 2006)[book
  reviews].
\newblock {\em IEEE Transactions on Neural Networks}, 20(3):542--542, 2009.

\bibitem{autoaug}
Ekin~D Cubuk, Barret Zoph, Dandelion Mane, Vijay Vasudevan, and Quoc~V Le.
\newblock Autoaugment: Learning augmentation policies from data.
\newblock {\em arXiv preprint arXiv:1805.09501}, 2018.

\bibitem{randaug}
Ekin~D Cubuk, Barret Zoph, Jonathon Shlens, and Quoc~V Le.
\newblock Randaugment: Practical automated data augmentation with a reduced
  search space.
\newblock In {\em Proceedings of the IEEE/CVF Conference on Computer Vision and
  Pattern Recognition Workshops}, pages 702--703, 2020.

\bibitem{kernelda}
Tri Dao, Albert Gu, Alexander~J Ratner, Virginia Smith, Christopher De~Sa, and
  Christopher R{\'e}.
\newblock A kernel theory of modern data augmentation.
\newblock {\em Proceedings of machine learning research}, 97:1528, 2019.

\bibitem{imagenet}
Jia Deng, Wei Dong, Richard Socher, Li-Jia Li, Kai Li, and Li Fei-Fei.
\newblock Imagenet: A large-scale hierarchical image database.
\newblock In {\em 2009 IEEE conference on computer vision and pattern
  recognition}, pages 248--255. Ieee, 2009.

\bibitem{cutout}
Terrance DeVries and Graham~W Taylor.
\newblock Improved regularization of convolutional neural networks with cutout.
\newblock {\em arXiv preprint arXiv:1708.04552}, 2017.

\bibitem{shakeshake}
Xavier Gastaldi.
\newblock Shake-shake regularization.
\newblock {\em arXiv preprint arXiv:1705.07485}, 2017.

\bibitem{timematters}
Aditya~Sharad Golatkar, Alessandro Achille, and Stefano Soatto.
\newblock Time matters in regularizing deep networks: Weight decay and data
  augmentation affect early learning dynamics, matter little near convergence.
\newblock In {\em Advances in Neural Information Processing Systems}, pages
  10678--10688, 2019.

\bibitem{affinity_diversity}
Raphael Gontijo-Lopes, Sylvia~J Smullin, Ekin~D Cubuk, and Ethan Dyer.
\newblock Affinity and diversity: Quantifying mechanisms of data augmentation.
\newblock {\em arXiv preprint arXiv:2002.08973}, 2020.

\bibitem{pyramidnet}
Dongyoon Han, Jiwhan Kim, and Junmo Kim.
\newblock Deep pyramidal residual networks.
\newblock In {\em Proceedings of the IEEE conference on computer vision and
  pattern recognition}, pages 5927--5935, 2017.

\bibitem{fasterautoaug}
Ryuichiro Hataya, Jan Zdenek, Kazuki Yoshizoe, and Hideki Nakayama.
\newblock Faster autoaugment: Learning augmentation strategies using
  backpropagation.
\newblock {\em arXiv preprint arXiv:1911.06987}, 2019.

\bibitem{he2016deep}
Kaiming He, Xiangyu Zhang, Shaoqing Ren, and Jian Sun.
\newblock Deep residual learning for image recognition.
\newblock In {\em Proceedings of the IEEE conference on computer vision and
  pattern recognition}, pages 770--778, 2016.

\bibitem{pba}
Daniel Ho, Eric Liang, Xi Chen, Ion Stoica, and Pieter Abbeel.
\newblock Population based augmentation: Efficient learning of augmentation
  policy schedules.
\newblock In {\em International Conference on Machine Learning}, pages
  2731--2741. PMLR, 2019.

\bibitem{huang2017densely}
Gao Huang, Zhuang Liu, Laurens Van Der~Maaten, and Kilian~Q Weinberger.
\newblock Densely connected convolutional networks.
\newblock In {\em Proceedings of the IEEE conference on computer vision and
  pattern recognition}, pages 4700--4708, 2017.

\bibitem{samplepairing}
Hiroshi Inoue.
\newblock Data augmentation by pairing samples for images classification.
\newblock {\em arXiv preprint arXiv:1801.02929}, 2018.

\bibitem{cifardataset}
Alex Krizhevsky, Geoffrey Hinton, et~al.
\newblock Learning multiple layers of features from tiny images.
\newblock 2009.

\bibitem{krizhevsky2012imagenet}
Alex Krizhevsky, Ilya Sutskever, and Geoffrey~E Hinton.
\newblock Imagenet classification with deep convolutional neural networks.
\newblock In {\em Advances in neural information processing systems}, pages
  1097--1105, 2012.

\bibitem{fastautoaug}
Sungbin Lim, Ildoo Kim, Taesup Kim, Chiheon Kim, and Sungwoong Kim.
\newblock Fast autoaugment.
\newblock In {\em Advances in Neural Information Processing Systems}, pages
  6665--6675, 2019.

\bibitem{svhn}
Yuval Netzer, Tao Wang, Adam Coates, Alessandro Bissacco, Bo Wu, and Andrew~Y
  Ng.
\newblock Reading digits in natural images with unsupervised feature learning.
\newblock 2011.

\bibitem{fixmatch}
Kihyuk Sohn, David Berthelot, Chun-Liang Li, Zizhao Zhang, Nicholas Carlini,
  Ekin~D Cubuk, Alex Kurakin, Han Zhang, and Colin Raffel.
\newblock Fixmatch: Simplifying semi-supervised learning with consistency and
  confidence.
\newblock {\em arXiv preprint arXiv:2001.07685}, 2020.

\bibitem{wu2020generalization}
Sen Wu, Hongyang~R Zhang, Gregory Valiant, and Christopher R{\'e}.
\newblock On the generalization effects of linear transformations in data
  augmentation.
\newblock {\em arXiv preprint arXiv:2005.00695}, 2020.

\bibitem{uda}
Qizhe Xie, Zihang Dai, Eduard Hovy, Minh-Thang Luong, and Quoc~V Le.
\newblock Unsupervised data augmentation for consistency training.
\newblock {\em arXiv preprint arXiv:1904.12848}, 2019.

\bibitem{shakedrop}
Yoshihiro Yamada, Masakazu Iwamura, Takuya Akiba, and Koichi Kise.
\newblock Shakedrop regularization for deep residual learning.
\newblock {\em IEEE Access}, 7:186126--186136, 2019.

\bibitem{wideresnet}
Sergey Zagoruyko and Nikos Komodakis.
\newblock Wide residual networks.
\newblock {\em arXiv preprint arXiv:1605.07146}, 2016.

\bibitem{mixup}
Hongyi Zhang, Moustapha Cisse, Yann~N Dauphin, and David Lopez-Paz.
\newblock mixup: Beyond empirical risk minimization.
\newblock {\em arXiv preprint arXiv:1710.09412}, 2017.

\bibitem{advaa}
Xinyu Zhang, Qiang Wang, Jian Zhang, and Zhao Zhong.
\newblock Adversarial autoaugment.
\newblock {\em arXiv preprint arXiv:1912.11188}, 2019.

\end{thebibliography}

\clearpage
\appendix

\appendixpage
\section{\kpp~Seeding Algorithm}
\label{app: kpp seeding}

As shown in Algorithm \ref{algo: kpp seeding}, the core idea of \kpp~seeding algorithm is to sample $S$ centers sequentially, where each new center is sampled with probability proportional to the squared distance to its nearest center.
The set of centers returned by Algorithm \ref{algo: kpp seeding} is theoretically guaranteed to far away from each others \cite{kppseeding}. 
\begin{algorithm}[hbt!]
\caption{\kpp seeding Algorithm~\cite{kppseeding}}
\label{algo: kpp seeding}
\KwIn{$G:=\{p_i:p_i\in\mathbb{R}^D\}$; Target size $S$}
\KwOut{Center set $C$ of size $S$}
$C_1=\{c_1\}$, where $c_1$ is sampled uniformly at random from $G$\\
\For{$t=2,\cdots,S$}{
    $E_t(x):=\min_{c\in C_{t-1}}||x-c||_2$\\
    $c_t\leftarrow$ sample $x$ from $G$ with probability $\frac{E_t^2(x)}{\sum_{x\in G}E_t^2(x)}$\\
    $C_t\leftarrow C_{t-1}\cup c_t$\\
}
\Return{$C_S$}
\end{algorithm}

\begin{table*}[hbt!]
\centering
\caption{Training hyperparameters of CIFAR-10, CIFAR-100 and ImageNet under the supervised settings. LR represents learning
rate, and WD represents weight decay. We do not specifically tune these hyperparameters, and all
hyperparameters are consistent with those reported in Adversarial AutoAugment \cite{advaa}.}
\label{tab: hp supervised}
\begin{tabular}{|cc|ccccc|} 
\hline
Dataset                                        & Model                                                                     & Batch Size & LR  & WD & Epoch & LR Schedule  \\ 
\hline
\multicolumn{1}{|l}{\multirow{4}{*}{CIFAR-10}} & \textcolor[rgb]{0.2,0.2,0.2}{Wide-ResNet-40-2}                            & 128        & 0.1 &  5e$-$4  & 200   & cosine       \\
\multicolumn{1}{|l}{}                          & \textcolor[rgb]{0.2,0.2,0.2}{Wide-ResNet-28-10}                           & 128        & 0.1 & 5e$-$4   & 200   & cosine       \\
\multicolumn{1}{|l}{}                          & \textcolor[rgb]{0.2,0.2,0.2}{Shake-Shake (26 2x96d)}                      & 128        & 0.2 &  1e$-$4  & 600   & cosine       \\
\multicolumn{1}{|l}{}                          & \multicolumn{1}{l|}{\textcolor[rgb]{0.2,0.2,0.2}{PyramidNet+ShakeDrop } } & 128        & 0.1 &  1e$-$4  & 600   & cosine       \\ 
\hline
\multirow{3}{*}{CIFAR-100}                     & \textcolor[rgb]{0.2,0.2,0.2}{Wide-ResNet-40-2}                            & 128       & 0.1 &  5e$-$4  & 200   & cosine       \\
                                              & \textcolor[rgb]{0.2,0.2,0.2}{Wide-ResNet-28-10}                           & 128       & 0.1 & 5e$-$4  & 200   & cosine       \\
                                              & \textcolor[rgb]{0.2,0.2,0.2}{Shake-Shake (26 2x96d)}                      & 128      & 0.1 &    5e$-$4 & 1200  & cosine       \\ 
\hline
ImageNet                                       & ResNet-50        & 512           & 0.2    & 1e$-$4   &   120    & cosine       \\
\hline
\end{tabular}
\end{table*}

\section{The details about the benchmark datasets}
\label{app: dataset stats}
The detailed statistic and the default data augmentation for the benchmark datasets are listed as belows.

\begin{itemize}
    \item \textbf{CIFAR-10 \& CIFAR-100 \cite{cifardataset}:} The training sets of the two datasets are composed of 50,000 colored images with 10 and 100 classes, respectively. Each image in these two datasets is in size of $32\times32$. 
    For CIFAR datasets, the default augmentation crops the padded image at a random location, and then horizontally flips it with the probability of 0.5.
    Then, it applies Cutout \cite{cutout} to randomly select a $16\times16$ patch of the image, and set the pixels within the selected patch as zeros.
    \item \textbf{SVHN \cite{svhn}}:
    This dataset contains color house-number images with 73,257 core images for training and 26,032 digits for testing.
    The default augmentation crops the padded image at a random location.
    Then it applies Cutout to randomly select a $16\times16$ patch of the image, and set the pixels within the selected patch as zeros.
    \item \textbf{ImageNet \cite{imagenet}:} ImageNet includes colored images of 1,000 classes. The training set has roughly 1.2M images, and the validation set has 50,000 images.
    The default augmentation randomly crops and resizes images to a size of $224\times224$, and then horizontally flips it with a probability of 0.5. Subsequently, it performs ColorJitter and PCA to the flipped image~\cite{krizhevsky2012imagenet}.
\end{itemize}

\section{Ablation Study}
\label{app: ablation study}

\subsection{Case Study}
In Figure \ref{fig: rotate_case_study}, DivAug’s  candidate  images  are  obtained by only applying the single transform \texttt{Rotate} with fixed probability parameter $p$ (the magnitude parameter remains random). As shown in Figure 6, Variance Diversity and model performance are highly correlated.

\begin{figure}[h!]
    \centering
    \includegraphics[scale=0.45]{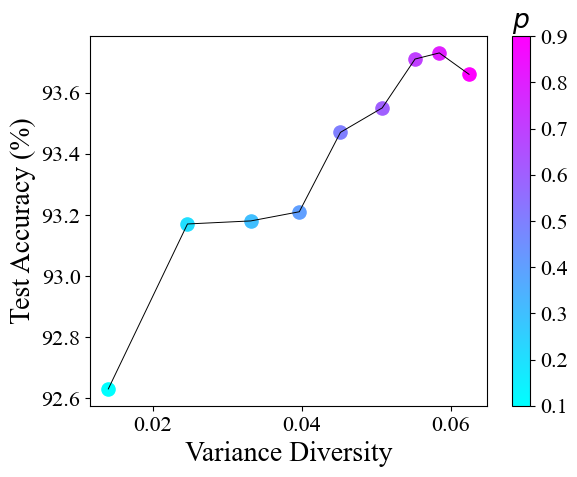}
    \caption{The case study of applying single transformation \texttt{Rotate}. To control Variance Diversity of augmented images, the fixed $p$ is varied from $0.1$ to $0.9$.}
    \label{fig: rotate_case_study}
\end{figure}

\subsection{Comparison between DivAug and the Random Baseline}

\begin{table}[h!]
\caption{The performance of Wide-ResNet-28-10 with DivAug and with the random baseline.}
\centering
\begin{tabular}{ccc} 
\hline
Dataset                   & Method & Accuracy  \\ 
\hline
\multirow{2}{*}{CIFAR10}  & Random ($S=4$) &  $97.7\pm.1$       \\
                          & DivAug & $98.1\pm.1$       \\ 
\hline
\multirow{2}{*}{CIFAR100} & Random ($S=4$) &    $83.3\pm.2$        \\
                          & DivAug & $84.2\pm.2$       \\
\hline
\end{tabular}

\label{tab: random_baseline}
\end{table}

To check the effect of \kpp~in DivAug, we compare the performance of Wide-ResNet-28-10 with DivAug and that with the random baseline on CIFAR-10 and CIFAR-100 in Table \ref{tab: random_baseline}.
For a fair comparison, the random baseline here randomly picks four augmented images from eight candidates for training.
Also, the magnitude $m$ and probability $p$ are also randomly picked.
As shown in Table \ref{tab: random_baseline}, DivAug is significantly better than the random baseline.

\section{Detailed Analysis For The Correlation between Variance Diversity and Generalization}
\label{app: detailed analysis of the correlation}
Recently, two measures, Affinity and Diversity, are introduced in \cite{affinity_diversity} for quantifying distribution shift and augmentation diversity, respectively. 
Across several benchmark datasets and models, it has been observed that the performance gain from data augmentation can be predicted not by either of these alone but by jointly optimizing the two \cite{affinity_diversity}.
Specifically, Affinity quantifies how much a sub-policy shifts the
training data distribution from the original one. 
For a set of augmented data, our proposed diversity measure is calculated based on the variance of their probability vectors. Meanwhile, the diversity measure proposed in \cite{affinity_diversity} is defined as the training loss of a given model over the augmented data.
Below, we give the formal definition of Affinity and Loss Diversity:

\begin{figure*}[t!]
\centering
    \begin{subfigure}[b]{.4\linewidth}
      \includegraphics[width=1\linewidth]{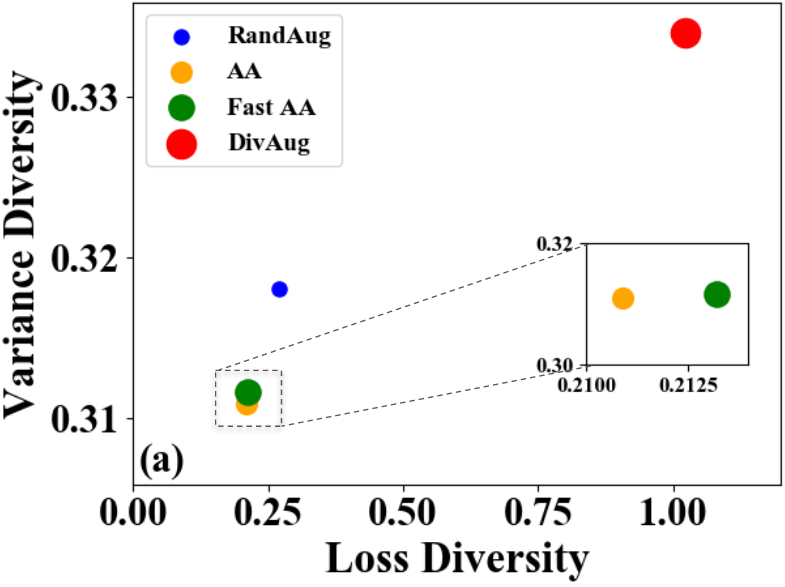}  
    \end{subfigure}
    \begin{subfigure}[b]{.4\linewidth}
    \includegraphics[width=1\linewidth]{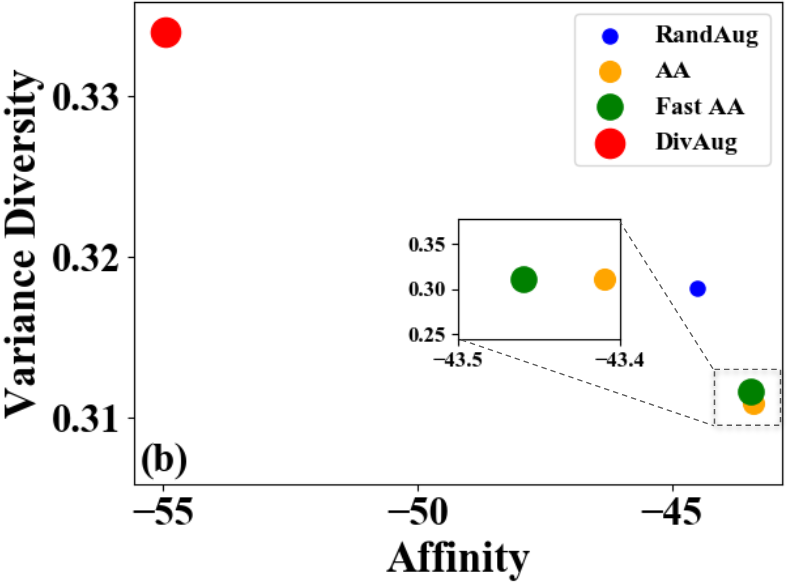}
    \end{subfigure}
    \caption{
    \textbf{
    The performance gain is positively correlated to Variance Diversity.
    Also, the Loss Diversity and Variance Diversity are highly correlated.}
    The marker size in the legend indicates the relative gain in test accuracy of different methods.
    (a) The Loss Diversity and the Variance Diversity of augmented data generated by different methods. All points lies near the diagonal of the Figure. In general, the relative gain in test accuracy increases with larger Variance Diversity 
    (b) The Affinity and Variance Diversity of augmented data generated by different methods.
    }
    \label{fig: var loss corr}
\end{figure*}

\begin{definition}[Affinity \cite{affinity_diversity}]
Let $D_{train}$ and $D_{val}$ be training and validation datasets drawn i.i.d. from the same clean data distribution, and let $D_{val}'$ be derived from $D_{val}$ by applying a stochastic augmentation strategy, $a$, once to each image in $D_{val}$, $D_{val}'=\{(a(x_i),y): \forall (x_i,y) \in D_{val}\}$.  
Further let $m$ be a model trained on $D_{train}$ and $\mathcal{A}(m, D)$ denote the model’s accuracy when evaluated on dataset D. The affinity $\tau[a;m;D_{val}]$ is defined as:
\begin{equation}
    \tau[a;m;D_{val}] = \mathcal{A}(m, D_{val}') - \mathcal{A}(m, D_{val})
\end{equation}
\end{definition}

\begin{definition}[Loss Diversity \cite{affinity_diversity}]
Let $D_{train}$ be the training set, and $D_{train}'$ be the augmented training set resulting from
applying a stochastic augmentation strategy $\alpha$.
For a set of augmented data $\mS=\{x_i'\}$, where $x_i'$ is obtained by applying $\alpha$ to $x_i$, stochastically.
Further, given a model $m$ which is trained on $D_{train}'$, let $\hat{L_i}$ be the training loss corresponding to $x_i'$. The Loss Diversity between $\{x_i'\}$, $\mathcal{D}_{\mathrm{loss}}(\{x_i'\})$, is defined as:
\begin{equation}
    \mathcal{D}_{\mathrm{loss}}(\mS)= \EXP_{x_i'\in \mS} \hat{L_i}~~\footnote{The original definition of Loss Diversity is defined for the entire training set. To make it comparable to Variance Diversity, we extend the concept to a set of augmented data generated from a same original data $x_i$.}
\end{equation}
\end{definition}
As we analyzed, given a set of augmented data which has large Variance Diversity, it is hard for models to give consist predictions for them, which will result in a large training loss. Thus, Loss Diversity and Variance Diversity are highly correlated.
The main difference between them is that Variance Diversity is a unsupervised measure, \ie, Variance Diversity is not related to the label information.

We further plot the performance gain from each augmentation methods against the Affinity, Loss Diversity, and Variance Diversity of the augmented data generated by them in Figure \ref{fig: var loss corr}.
In the legend, the marker size indicates the test accuracy of a Wide-ResNet-40-2 model trained with different automated data augmentation methods (The detailed results are shown in the first row of Table \ref{tab: supervised res}).
Figure \ref{fig: var loss corr} demonstrates the Loss Diversity and Variance Diversity are highly correlated, which is consistent with our theoretical analysis.
Following \cite{affinity_diversity}, we show the Affinity and Variance Diversity of augmented data generated by different methods in Figure \ref{fig: var loss corr} (b).
There is a clear trend that the Loss Diversity and Variance Diversity contradict with the Affinity to some extent.
We remark that although RA has larger Variance Diversity than AA and Fast AA, the performance gain from RA is smaller.
According to the hypothesis in \cite{affinity_diversity}, this can be explained by RA has smaller Affinity than those of AA and Fast AA.
In contrast, although DivAug has the largest Variance Diversity, largest Loss Diversity, and the smallest Affinity, DivAug performs best in terms of the test accuracy.
We hypothesize that there might exist a sweet spot between the Diversity and Affinity, and how to achieve this sweet spot is a interesting future direction for the automated data augmentation methods.

\section{Experiment Details}
\label{app: experiment details}
We list the details of training hyperparameters from the experiments in Section \ref{exp: supversied} in Table \ref{tab: hp supervised}.

For the semi-supervised learning experiment in Section \ref{exp: ssl},  
we follow the settings in \cite{uda} and employ Wide-ResNet-28-
2 \cite{wideresnet} as the backbone model and evaluate UDA \cite{uda} with varied supervised data sizes.
For the experiments on CIFAR-10 with supervised data size 1000, 2000, and 4000, the hyperparameters of them are identical as below: 
we train the backbone model for 200K steps.
We use a batch size of 32 for labeled data and a batch size of 448 for unlabeled data.
The softmax temperature $\tau$ is set to 0.4.
The confidence threshold $\beta$ is set to 0.8.
The backbone model is trained by a SGD optimizer with learning rate of 1e$-$4, weight decay of 5e$-$4, and 
the nesterov momentum with the momentum hyperparameter set to 0.9.
We remark that all hyperparameters are identical to those reported in \cite{uda}, except two differences: we train the backbone model for 200K steps instead of 500K, and we do not apply Exponential Moving Average to the parameters of backbone model.

\end{document}